\documentclass[runningheads]{llncs}
\usepackage[T1]{fontenc}
\usepackage{graphicx,verbatim}
\usepackage{amssymb}
\usepackage{booktabs}
\usepackage{multirow}
\usepackage{arydshln} 
\usepackage{pifont}
\usepackage{makecell}
\usepackage{amsmath}
\usepackage{hyperref}
\usepackage[table,xcdraw]{xcolor}
\usepackage{marvosym}
\begin{document}
\title{Propagating Structural Guidance: Synthesizing Fluorescein Angiography from Fundus Images and Sparse OCT Scans
}
%\titlerunning{Abbreviated paper title}
% If the paper title is too long for the running head, you can set
% an abbreviated paper title here
%
\begin{comment}  %% Removed for anonymized MICCAI submission
\author{First Author\inst{1}\orcidID{0000-1111-2222-3333} \and
Second Author\inst{2,3}\orcidID{1111-2222-3333-4444} \and
Third Author\inst{3}\orcidID{2222--3333-4444-5555}}
%
\authorrunning{F. Author et al.}
% First names are abbreviated in the running head.
% If there are more than two authors, 'et al.' is used.
%
\institute{Princeton University, Princeton NJ 08544, USA \and
Springer Heidelberg, Tiergartenstr. 17, 69121 Heidelberg, Germany
\email{lncs@springer.com}\\
\url{http://www.springer.com/gp/computer-science/lncs} \and
ABC Institute, Rupert-Karls-University Heidelberg, Heidelberg, Germany\\
\email{\{abc,lncs\}@uni-heidelberg.de}}

\end{comment}

% \author{Anonymized Authors}  %% Added for anonymized MICCAI submission
% \authorrunning{Anonymized Author et al.}
% \institute{Anonymized Affiliations \\
%     \email{email@anonymized.com}}
\author{
    Tengfei Ma\inst{1,2}, 							% index{Ma Tengfei}
    Ruiqi Wu\inst{1,2},								% index{Wu Ruiqi}
    Chenran Zhang\inst{1,2},						% index{Zhang Chenran}
    Ye Geng\inst{3},								% index{Geng Ye}
    Na Su\inst{5}, \\								% index{Su Na}
    Xiangyuan Duanmu\inst{3},						% index{Duanmu Xiangyuan}
    Tao Zhou\inst{4},								% index{Zhou Tao}
    Yi Zhou\inst{1,2}\textsuperscript{(\Letter)},	% index{Zhou Yi}
    Wen Fan\inst{5}\textsuperscript{(\Letter)}		% index{Fan Wen}
}
\authorrunning{Ma et al.}
\institute{
    School of Computer Science and Engineering, Southeast University, China \inst{1} \\
    Key Laboratory of New Generation Artificial Intelligence Technology and \\ Its Interdisciplinary Applications, Ministry of Education, Nanjing, China \inst{2} \\
    Tianyuan Honors School, Nanjing Medical University, China \inst{3} \\
    Nanjing University of Science and Technology, Nanjing, China \inst{4} \\
    Department of Ophthalmology, The First Affiliated Hospital of \\ Nanjing Medical University, Nanjing, China \inst{5} \\
    \email{\{matengfei\_1013@163.com, yizhou.szcn@gmail.com\}}
}
\maketitle              % typeset the header of the contribution
\begin{abstract}
Fundus fluorescein angiography (FFA) is critical for assessing retinal vascular abnormalities, but its acquisition is invasive and not always feasible. In contrast, color fundus photography (CFP) is non-invasive and widely accessible, which has motivated studies on CFP-to-FFA synthesis. However, prior works rely solely on CFP surface texture, fundamentally limiting the ability to reconstruct functional vascular information and subtle pathological changes.
To address this, we propose a novel framework that synthesizes FFA from CFP with structural guidance provided by optical coherence tomography (OCT).
We construct a multi-modal retinal imaging dataset with paired CFP, FFA, and OCT from 3,676 patient eyes—the first tri-modally aligned dataset in retinal imaging.
To bridge the spatial gap between OCT and fundus modalities, we propose a Spatially Aligned Cross-Modal Fusion (\textbf{SACMF}) module that projects depth-resolved OCT features onto the fundus plane and injects them into the CFP encoder via adaptive layer normalization.
Beyond feature fusion, we further introduce Token-wise Cross-Modality Alignment (\textbf{TCMA}), a token-level contrastive learning strategy that explicitly aligns CFP and FFA representations at corresponding spatial positions.
Our method achieves superior synthesis performance compared to state-of-the-art methods.
Moreover, extensive experiments demonstrate that the FFA images synthesized by our approach bring greater improvements in downstream disease diagnosis performance than existing methods, highlighting the clinical potential of our approach as a non-invasive decision-support tool in routine workflows. The code is available at 
\href{https://github.com/while-plus/OCT-guide-FFA-Syn}{https://github.com/while-plus/OCT-guide-FFA-Syn}.
% \href{https://anonymous.4open.science/r/OCT-guide-FFA-Syn-D6CE}{here}.

\end{abstract}
\section{Introduction}
% Fundus fluorescein angiography (FFA) plays a critical role in retinal disease assessment by visualizing vascular perfusion, leakage, and pathological microcirculation changes. These vascular cues are essential for diagnosing and monitoring diseases such as diabetic retinopathy and retinal vein occlusion \cite{kylstra1999importance}. However, FFA is an invasive procedure requiring intravenous dye injection, limiting its routine and repeated use.
% In contrast, color fundus photography (CFP) is non-invasive and easy to acquire in routine clinical practice, making it suitable for large-scale screening and longitudinal follow-up.
% Nevertheless, CFP primarily captures retinal surface appearance and color contrast, lacking explicit information about vascular permeability and dynamic blood flow \cite{kylstra1999importance}. This fundamental limitation restricts its ability to directly reflect functional vascular abnormalities that are central to many retinal pathologies. This gap motivates the exploration of computational approaches that infer vascular-related information from CFP, aiming to approximate FFA without direct angiographic acquisition.
	
Fundus fluorescein angiography (FFA) plays a critical role in retinal disease assessment by visualizing vascular perfusion, leakage, and pathological microcirculation changes, essential for diagnosing diabetic retinopathy and retinal vein occlusion \cite{kylstra1999importance}. However, FFA is invasive, requiring intravenous dye injection, which limits routine and repeated use. In contrast, color fundus photography (CFP) is non-invasive and easily acquired \cite{yannuzzi2004ophthalmic}, suitable for large-scale screening and longitudinal follow-up. Nevertheless, CFP primarily captures retinal surface appearance, lacking explicit information about vascular permeability and dynamic blood flow \cite{kylstra1999importance}. This limitation restricts its ability to reflect functional vascular abnormalities central to many retinal pathologies. This gap motivates computational approaches that infer vascular-related information from CFP, aiming to approximate FFA without direct angiographic acquisition.

Despite recent advances in generative modeling for FFA synthesis, most existing works \cite{wang2024non,huang2023lesion,9412428,kamran2020fundus2angio} rely solely on CFP input. Consequently, these approaches are constrained by CFP's limited information, as it does not encode vascular leakage or depth-resolved tissue properties. This fundamentally restricts the upper bound of CFP-to-FFA synthesis, often producing images lacking fine-grained vascular details. More critically, networks may generate fluorescence patterns based on superficial texture correlations rather than true vascular structures. Such artifacts underscore the intrinsic limitations of CFP-only approaches and severely constrain the clinical interpretability of synthesized FFA.

To address these limitations, our study is the first to incorporate optical coherence tomography (OCT) as complementary structural guidance for FFA generation, bridging the modality gap between fundus photography and angiography. Similar to CFP, OCT is non-invasive and safely acquired when FFA is contraindicated, making it a practical auxiliary modality. In intraoperative scenarios, only CFP-like views and OCT are typically available, as FFA is invasive and procedurally impractical. Therefore, synthesizing FFA from CFP and OCT is valuable for surgical decision-making requiring real-time vascular assessment \cite{ehlers2014value}.
Importantly, OCT provides high-resolution, layer-resolved structural information that CFP cannot capture. Structural abnormalities such as retinal thickening or abnormal reflectivity patterns in OCT are often associated with fluorescence leakage or accumulation in FFA \cite{liu2017pathophysiological}. Incorporating OCT introduces anatomically grounded cues, reducing the gap between surface appearance and underlying vascular pathology, enabling more reliable fluorescence pattern modeling.
Considering clinical workflows, only limited representative two-dimensional B-scan slices are typically retained instead of full volumetric scans. Accordingly, we focus on sparse B-scans as input, ensuring practical feasibility while reducing data acquisition burden and storage requirements.
    
\textbf{Our major contributions are highlighted as:} \textbf{1)} We are the first to investigate CFP-to-FFA synthesis in a multi-modal pipeline by incorporating OCT as auxiliary structural guidance, constructing the first paired tri-modal retinal dataset with 3,676 aligned CFP, FFA, and OCT cases. \textbf{2)} We propose an OCT-guided feature modulation module that projects depth-aware OCT representations onto the fundus plane and injects them into the CFP encoder via adaptive normalization. \textbf{3)} We introduce a token-wise cross-modality contrastive learning strategy that enforces CFP-FFA alignment at matched spatial positions. \textbf{4)} Extensive experiments demonstrate that our approach generates FFA images with higher visual fidelity, achieving superior downstream diagnostic performance, and exhibits greater clinical utility compared to baselines.
% produces more discriminative representations for downstream disease diagnosis tasks.

%, while preserving asymmetry to prevent redundancy, thereby improving local vascular fidelity and cross-modality consistency.
\begin{figure}[t]
    \centering
    \includegraphics[scale=0.34]{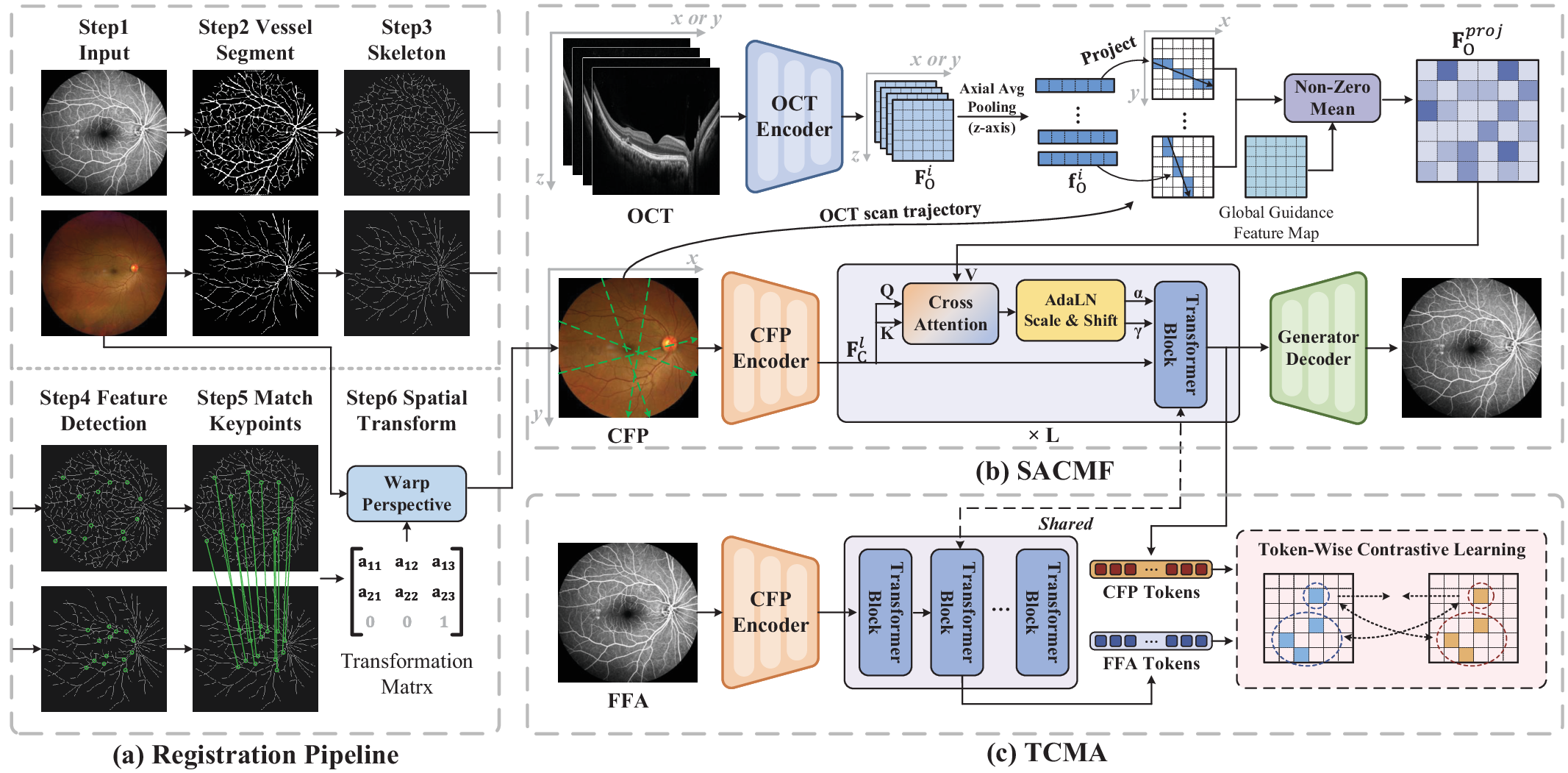}
    \caption{Registration Pipeline and Overall Architecture. 
    (a) Registration pipeline. Establishes spatial correspondence between CFP and FFA.
    (b) SACMF. Bridges the viewpoint gap between OCT and CFP and propagates structural guidance across the fundus plane. Green dashed lines indicate OCT scan trajectories.
    (c) TCMA. Enforces fine-grained semantic consistency between CFP and FFA.
    }

    \label{fig:1}
\end{figure}
		% (a): Registration pipeline. A paired FFA and CFP are processed through vessel segmentation, skeleton extraction, keypoint matching, and perspective warping to establish spatial alignment. 
		% (b) SACMF. OCT B-scans are projected onto the en-face plane to form a spatially aligned structural map, which modulates CFP features through cross-attention and AdaLN. The green lines overlaid on the CFP indicate the OCT scan trajectories.
		% (c) TCMA. Token-wise contrastive module further aligns CFP and FFA representations.
\section{Methods}
\subsection{Overview}
\subsubsection{Data Curation.}
%% 数据集处理与构建
Each case was collected from real-world clinical practice at a hospital and includes one CFP, one FFA, and multiple clinician-selected OCT slices considered diagnostically informative. As illustrated in Fig. \ref{fig:1}(a), CFP and FFA are first spatially registered to enable cross-modal correspondence. 
Vessel segmentation is performed for both modalities. CFP vessels are segmented using Automorph \cite{zhou2022automorph}, while FFA vessels are segmented using a model trained with a style loss \cite{zhang2019joint} on the CHASE \cite{owen2011retinal}, DRIVE \cite{1282003}, HRF \cite{budai2013robust} and STARE \cite{845178} datasets. From the segmented vessels, skeletons \cite{van2014scikit} are extracted to identify endpoints and bifurcation points. ORB descriptors \cite{rublee2011orb} are computed at these keypoints to characterize local vessel structures.
Feature correspondences are then matched to estimate a homography matrix. The estimated transformation is applied via perspective warping to align both the vessel masks and the fundus images. Cases with a Dice coefficient \cite{milletari2016v} below 0.4 between the aligned vessel masks are excluded. Thus, our final curated dataset contains 3,676 cases, with an average of 2.88 OCT slices per case.

\subsubsection{Overall Architecture.}
Although diffusion models \cite{ho2020denoising} have recently achieved remarkable performance in natural image generation, our task involves paired cross-modal translation with limited medical data and strict structural consistency constraints. 
In this scenario, conditional GANs show promise due to their strong pixel-level supervision and training efficiency. 
Therefore, we adopt a Generative Adversarial Network (GAN) \cite{kong2021breaking} as the backbone for CFP-to-FFA synthesis. 
The framework consists of three main components: (1) a generator with dual encoders for CFP and OCT feature extraction, (2) the proposed Spatial Aligned Cross-Modal Fusion (SACMF, Fig. \ref{fig:1}(b)) module for structurally consistent multi-modal fusion, (3) a Token-wise Cross-Modality Alignment (TCMA, Fig. \ref{fig:1}(c)) module that enforces fine-grained semantic consistency during training. A discriminator is further introduced to encourage the realism of the synthesized FFA images under adversarial supervision.
% As illustrated in Fig. \ref{fig:1}(b), we introduce the Spatial Aligned Cross-Modal Fusion (SACMF) module to bridge the structural disparity between OCT and CFP. Specifically, the generator contains two parallel encoders that process CFP and OCT, respectively. The OCT branch takes multiple B-scan slices as input and feeds them into a frozen pretrained OCT encoder \cite{wu2025mm}, yielding 1D feature vectors that encode deep retinal structural information. Guided by the scans’ spatial coordinates, these features are projected onto the 2D fundus plane, forming a sparse feature map aligned with the CFP geometry. This map is then fused with CFP features via adaptive layer normalization (AdaLN) modulation at multiple layers.
% In addition, as shown in Fig. \ref{fig:1}(c), we propose a Token-wise Cross-Modality Alignment (TCMA) module to further enhance fine-grained semantic consistency between CFP and FFA representations. During training, FFA images are passed through the same CFP encoder to obtain their feature representations. Features at corresponding spatial locations between CFP and FFA are treated as positive pairs. In contrast, non-corresponding tokens are regarded as negative samples to enforce fine-grained cross-modality alignment.

\subsection{Spatial Aligned Cross-Modal Fusion (SACMF)}
%Due to the inherent viewpoint discrepancy between OCT B-scans, which capture cross-sectional retinal anatomy along the $x-z$ or $y-z$ plane, and CFP/FFA, which represent en-face fundus appearance in the $x-y$ plane, directly fusing features from the two modalities tends to introduce significant spatial inconsistency.
% To address this issue, the proposed module aims to preserve OCT structural information while constructing a spatially consistent representation aligned with the 2D fundus plane.

The proposed SACMF module consists of two sequential steps: 
(1) trajectory-aware en-face projection and 
(2) CFP-guided structural propagation. 
The first step aims to resolve the viewpoint discrepancy between cross-sectional OCT B-scans and en-face fundus images by constructing a spatially aligned structural representation on the 2D fundus plane. 
The second step propagates sparse OCT-derived structural cues to CFP regions without direct OCT coverage by leveraging appearance similarity among CFP feature tokens, thereby enabling global structural guidance for FFA synthesis.

\textbf{Trajectory-Aware En-face Projection.}
Due to the viewpoint discrepancy between OCT B-scans, which capture retinal anatomy along the $x$-$z$ or $y$-$z$ plane \cite{li2024octa}, and CFP/FFA, which represent en-face appearance in the $x$-$y$ plane, directly fusing the two modalities introduces spatial inconsistency. 

Given $N$ OCT B-scan slices $\mathbf{X}_\mathrm{O}^{i}$ $(i=1\cdots N)$ co-registered with the fundus space, each slice is processed by a frozen pretrained OCT encoder $ \mathbf{E}_\mathrm{O} $, producing feature maps $\mathbf{F}_{\mathrm{O}}^i \in \mathbb{R}^{C \times H \times W}$. 
Average pooling is applied along the axial ($z$) direction to obtain strip-like representations $ \mathbf{f}_{\mathrm{O}}^i \in \mathbb{R}^{C \times 1 \times W}$. 
According to the known scanning trajectories, each $ \mathbf{f}_{\mathrm{O}}^i $ is embedded into a zero-initialized 2D feature map $ \mathbf{M}_i \in \mathbb{R}^{C \times H \times W}$ at its corresponding location. 
All embedded maps $\{\mathbf{M}_i\}_{i=1}^N$ are aggregated with a learnable global guidance map $ \mathbf{G} \in \mathbb{R}^{C \times H \times W}$ via non-zero-aware averaging:
\begin{equation}
    \mathbf{F}_{\mathrm{O}}^{proj} =
    \frac{\sum_{i=1}^{N} \mathbf{M}_i + \mathbf{G}}
    {\sum_{i=1}^{N} \mathbb{I}[\mathbf{M}_i \neq \mathbf{0}] + \mathbb{I}[\mathbf{G} \neq \mathbf{0}]},
\end{equation}
where $ \mathbb{I}[\cdot] $ denotes an element-wise indicator function.

\textbf{CFP-Guided Structural Propagation.}
Although $\mathbf{F}_{\mathrm{O}}^{proj}$ provides spatially aligned structural anchors, large regions in CFP remain without direct OCT measurements. 
To propagate structural cues beyond sparse trajectories, we exploit long-range appearance similarity within CFP features.
Given the CFP input $\mathbf{X}_\mathrm{C}$, the $l$-th encoder layer produces $\mathbf{F}_\mathrm{C}^l$. 
The query and key are derived from CFP features to measure appearance compatibility, while the projected OCT feature serves as the value to inject structural information. 
Accordingly, structural guidance is propagated as follows:
\begin{equation}
    \mathbf{F}_{\mathrm{prop}}^l  =
    \mathrm{softmax}\left(
    \frac{Q(\mathbf{F}_{\mathrm{C}}^l)K(\mathbf{F}_{\mathrm{C}}^l)^\top}{\sqrt d}
    \right) V(\mathbf{F}_{\mathrm{O}}^{proj}).
\end{equation}
This allows OCT-uncovered pixels to retrieve structural cues via CFP-based similarity.
The propagated feature $\mathbf{F}_{\mathrm{prop}}^l$ is injected into $\mathbf{F}_\mathrm{C}^l$ via adaptive layer normalization for structure-aware FFA synthesis.

\subsection{Token-wise Cross-Modality Alignment (TCMA)}
% In the second module, feature-level supervision from FFA is used to guide the CFP branch to learn functionally consistent representations. For each input CFP–FFA pair, the shared CFP encoder $ \mathbf{E}_\mathrm{C} $ is used to generate multi-layer token representations $ \mathbf{T}_\mathrm{C}\in\mathbb{R}^{K\times d}$ and $ \mathbf{T}_\mathrm{F}\in\mathbb{R}^{K\times d}$, where K is the number of spatial tokens and $d$ is the feature dimension.

For each input CFP–FFA pair, both images are first processed by a shared CFP encoder, followed by a stack of $L$ transformer blocks. 
% The output of the $l$-th transformer block is taken as the token representation, denoted as $\mathbf{T}_\mathrm{C} \in \mathbb{R}^{K \times d}$ for CFP and $\mathbf{T}_\mathrm{F} \in \mathbb{R}^{K \times d}$ for FFA, where $K$ is the number of spatial tokens and $d$ is the feature dimension. 
The $l$-th block outputs CFP and FFA token representations 
$\mathbf{T}_\mathrm{C}, \mathbf{T}_\mathrm{F} \in \mathbb{R}^{K \times d}$, where $K$ and $d$ denote the number of spatial tokens and feature dimension.
Notably, the CFP tokens $\mathbf{T}_\mathrm{C}$ are adaptively modulated using OCT-derived information via AdaLN, while the FFA tokens $\mathbf{T}_\mathrm{F}$ remain unmodulated.
This asymmetric design allows CFP tokens, conditioned on OCT guidance, to consistently reflect the underlying FFA signals, while keeping FFA tokens unmodulated to prevent redundant or biased information.
To further promote cross-modality consistency, we adopt a symmetric InfoNCE objective to enforce bidirectional alignment between CFP and FFA tokens. Tokens at the same spatial location are treated as positive pairs, and all other tokens as negatives. Specifically, the InfoNCE loss is defined as: %\cite{he2020momentum} 
\begin{equation}
    \mathcal{L}_{\mathrm{InfoNCE}}(\mathbf{q}, \mathbf{k}^+, \{\mathbf{k}^j\}_{j=1}^{K}) = 
    -\log
    \frac{\exp(\mathrm{sim}(\mathbf{q},\mathbf{k}^+)/\tau)}
    {\sum_{j=1}^{K}\exp(\mathrm{sim}(\mathbf{q},\mathbf{k}^j)/\tau)}, 
\end{equation}
where $ \mathbf{q} $ is a query token, $ \mathbf{k}^+ $ is its positive key, and $ \{\mathbf{k}^j\}_{j=1}^{K} $ denotes the set of keys for contrastive comparison. $ \mathrm{sim}(\cdot,\cdot) $ denotes cosine similarity and $ \tau $ is a temperature hyperparameter.
Accordingly, the token-wise contrastive loss is formulated as:
\begin{equation}
    \begin{split}
        \mathcal{L}_{\mathrm{tok\_cont}}(T_C, T_F) = \frac{1}{K}\sum_{i=1}^{K} \Big( 
        &\mathcal{L}_{\mathrm{InfoNCE}}(\mathbf{T}_\mathrm{C}^i,\mathbf{T}_\mathrm{F}^i, \{\mathbf{T}_\mathrm{F}^j\}) \\
        &+ \mathcal{L}_{\mathrm{InfoNCE}}(\mathbf{T}_\mathrm{F}^i,\mathbf{T}_\mathrm{C}^i,\{\mathbf{T}_\mathrm{C}^j\}) \Big).
    \end{split}
\end{equation}
Applying this supervision across multiple layers aligns structural and functional representations while preserving local details in the fundus image.
Notably, TCMA is activated only during training with real FFA tokens for contrastive supervision; during inference, the model relies solely on CFP inputs.

\subsubsection{Overall Training Objective.}
In addition to the proposed token-wise contrastive loss, we follow the baseline setting in \cite{kong2021breaking}, incorporating registration loss $\mathcal{L}_{\mathrm{Reg}}$ and adversarial loss $\mathcal{L}_{\mathrm{GAN}}$ to ensure stable training and realistic synthesis.
Consequently, the final objective is defined as:
\begin{equation}
    L = \mathcal{L}_{\mathrm{GAN}} + \mathcal{L}_{\mathrm{Reg}} + \beta \mathcal{L}_{\mathrm{tok\_cont}}, 
    \label{final obj}
\end{equation}
where $\beta$ balances the contribution of the contrastive supervision.

% The token-wise contrastive loss is defined as
%$$\mathcal{L}_{contrast}=-\frac{1}{N}\sum_{i=1}^{N}{\log}\frac{\exp(sim(\mathbf{T}_\mathrm{C}^i,\mathbf{T}_\mathrm{F}^i)/\tau)}{\sum_{j=1}^{N}{\exp(}sim(\mathbf{T}_\mathrm{C}^i, \mathbf{T}_\mathrm{F}^j)/\tau)} + {\log} \frac{\exp(sim(\mathbf{T}_\mathrm{C}^i,\mathbf{T}_\mathrm{F}^i)/\tau)}{\sum_{j=1}^{N}{\exp(}sim(\mathbf{T}_\mathrm{C}^j, \mathbf{T}_\mathrm{F}^i)/\tau)}$$
%where $sim(\cdot,\cdot)$ denotes cosine similarity and $\tau$ is a temperature hyperparameter.
% 补充介绍最终的所有损失 + Adversial Loss

\section{Experiments}
\subsubsection{Implementation Details.}
All images were resized to $512 \times 512$. The synthesis model trained for 160 epochs using Adam with a learning rate of 1e-5 and a batch size of 4. The dataset split was 60\% training and 40\% testing. The weighting coefficient $\beta$ in Eq. \ref{final obj} was set to 0.1. All experiments were conducted on four NVIDIA GeForce RTX 4090 GPUs.
	
To evaluate diagnostic utility, we performed a downstream classification task on three prevalent diseases: diabetic retinopathy, retinitis, and choroidal disorders. Using 75\% of these samples for training and the rest for testing, we employed a ResNet18 backbone optimized by Adam at a learning rate of 5e-4 for 50 epochs. This setup assesses whether synthesized images enhance disease diagnosis performance beyond standard image quality metrics.
% All images were resized to $512 \times 512$ pixels. The model was trained for 160 epochs using the Adam optimizer with a learning rate of 1e-5 and a batch size of 4. Random horizontal flipping was applied for data augmentation. The dataset was split into 60\% for training and 40\% for testing. All experiments were conducted on four NVIDIA GeForce RTX 4090 GPUs. The weighting coefficient $\beta$ in Eq. \ref{final obj} was empirically set to 0.1.

% In addition to image synthesis evaluation, we further assessed whether the synthesized images improve disease diagnosis via a downstream classification task. From the test set, we selected the three most prevalent diseases: diabetic retinopathy, retinitis, and choroidal disorders. Among these samples, 75\% were used for training and the remaining for testing. ResNet18 was adopted as the backbone and optimized with Adam at a learning rate of 5e-4 for 50 epochs. 
% In addition, the image encoder $f(\cdot)$ of a CLIP model was trained on the FFA-IR dataset to measure the similarity between synthesized and real images. The training protocol follows Chinese CLIP. The CLIP Score is computed as the cosine similarity between the normalized embeddings of the synthesized image $\hat{I}$ and the real image $I$: 
% $
% \mathrm{CLIP\text{-}Score}(\hat{I}, I) =
% \frac{ f(\hat{I}) \cdot f(I) }
% { | f(\hat{I}) |_2 , | f(I) |_2 }
% $

%	where $f(\cdot)$ denotes the image encoder.

\subsubsection{Evaluation Criteria.}
For image quality evaluation, we report PSNR \cite{huynh2008psnr}, SSIM \cite{wang2004ssim}, LPIPS \cite{zhang2018lpips}, and FID \cite{heusel2017fid} to assess pixel fidelity, structural consistency, perceptual similarity, and distribution alignment. 
Classification performance is measured by Precision (Pre), Recall (Rec), and F1-score (F1).
% For the classification task, Precision (Pre), Recall (Rec), and F1-score (F1) are adopted to measure diagnostic performance. 
The lower bound takes OCT and CFP as input, while the upper bound additionally incorporates real FFA. All other methods follow the upper-bound setting but replace real FFA with synthesized FFA for classification. % In addition, CLIP Score is used to evaluate the semantic consistency between synthesized and real FFA images. 
Semantic consistency is evaluated via cosine similarity (Cos Sim) between feature embeddings of synthesized and real images, extracted by an encoder trained on FFA-IR \cite{li2021ffa} \cite{hessel2021clipscore}.
% To quantitatively evaluate the semantic consistency \cite{hessel2021clipscore} between synthesized and real images, we compute the cosine similarity (Cos Sim) between their feature embeddings extracted by an image encoder trained on the FFA-IR \cite{li2021ffa} dataset.

\begin{table}[t]
    {\small
        \centering
        \caption{Comparison of different methods. The best results are highlighted in \textbf{bold}.} %The lower bound uses OCT and CFP as input, while the upper bound uses OCT, CFP, and real FFA as input.}
        \label{tab:comp}
        \resizebox{0.98\linewidth}{!}{
            \begin{tabular}{c cccc cccc
                }
                % c @{\hskip 3mm} cccc @{\hskip 3mm} cccc @{\hskip 2.5mm
                \toprule
                \toprule
                \multirow{2}{*}{\bf Method} & \multicolumn{4}{c}{Image Quality} & \multicolumn{3}{c}{Diagnosis(\%)} & \multicolumn{1}{c}{Semantic(\%)}  \\ % \multirow{2}{*}{\thead{Clip \\ Score$\uparrow$}} 
                \cmidrule(r){2-5} \cmidrule(r){6-8} \cmidrule(r){9-9} 
                
                & PSNR$\uparrow$ & SSIM$\uparrow$ & FID$\downarrow$ & LPIPS$\downarrow$ & Pre$\uparrow$ & Rec$\uparrow$ & F1$\uparrow$ & Cos Sim$\uparrow$\\
                \midrule
                Lower bound & - & - & - & - & 45.59 & 42.66 & 43.16 & - \\
                Upper bound & - & - & - & - & 54.18 & 53.66 & 53.80 & - \\
                \midrule
                AAGAN \cite{9412428} & 15.48 & 0.4370 & 11.23 & 0.2224 & 47.40 & 45.61 & 46.23 & 76.59 \\
                BBDM \cite{li2023bbdm} & 13.77 & 0.3167 & 16.23 & 0.2816 & 45.28 & 44.32 & 44.57 & 59.01 \\
                FAGAN \cite{kamran2020fundus2angio} & 15.54 & 0.4276 & 10.22 & 0.2146 & 48.11 & 47.68 & 47.76 & 77.30 \\
                P2P \cite{isola2017image} & 15.22 & 0.4316 & 13.92 & 0.2294 & 48.89 & 46.31 & 47.16 & 77.84 \\
                ResVit \cite{dalmaz2022resvit} & 15.59 & 0.4325 & 12.89 & 0.2172 & 48.96 & 47.30 & 47.97 & 77.28 \\
                RegGAN \cite{kong2021breaking} & 15.36 & 0.4223 & 12.74 & 0.2135 & 46.58 & 45.44 & 45.84 & 76.58 \\
                \rowcolor[HTML]{D9D9D9}\textbf{Ours} & \bf{15.87} & \bf{0.4502} & \bf{10.05} & \bf{0.2009} & \bf{52.83} & \bf{50.61} & \bf{51.49} & \bf{79.13} \\

                %\cellcolor[HTML]{D9D9D9}
                \toprule
                \toprule
            \end{tabular}
        }
        
        \vspace{1mm}
        \vspace{-5mm}
        %		\lbltbl{baselines}
    }
\end{table}

\subsubsection{Quantitative Comparisons.}
Table \ref{tab:comp} reports quantitative comparisons with existing fundus translation methods \cite{9412428,kamran2020fundus2angio} and representative natural and medical image generation models \cite{li2023bbdm,isola2017image,dalmaz2022resvit,kong2021breaking}.
\textbf{Image Quality.}
Our method achieves the best performance across PSNR, SSIM, FID, and LPIPS, demonstrating superior structural fidelity and perceptual realism. The consistent gains over both fundus-specific and general generation models validate the benefit of incorporating OCT as structural guidance for angiographic synthesis.
\textbf{Diagnosis \& Semantic Consistency.}
It also shows that using synthesized FFA improves the lower bound of multi-disease classification. 
% Our approach attains the highest Precision, Recall, and F1-score among all synthesis methods, narrowing the gap toward the upper bound with real FFA. 
Our approach attains the highest Precision, Recall, and F1-score, narrowing the gap toward the upper bound with real FFA.
Moreover, it achieves the highest cosine similarity in semantic embedding space, indicating stronger alignment with real angiographic representations.
% \textbf{Image Quality.}
% % Our method achieves the best performance across all image quality metrics, reaching 15.87 PSNR and 0.4502 SSIM, while reducing FID to 10.05. Notably, compared with the strongest baseline FAGAN (FID 10.22), our model further improves perceptual realism and structural fidelity. These gains validate the effectiveness of OCT-guided structural modeling.
% \textbf{Diagnosis \& Semantic Consistency.}
% Using synthesized FFA significantly improves the lower-bound classification performance (F1: 43.16\% → 51.49\%), narrowing the gap toward the upper bound with real FFA (53.80\%). Our method also achieves the highest cosine similarity (79.13\%), indicating stronger semantic alignment with real angiographic representations.

\subsubsection{Qualitative Comparison.}
\begin{figure}[t!]
    \centering
    \includegraphics[scale=0.38]{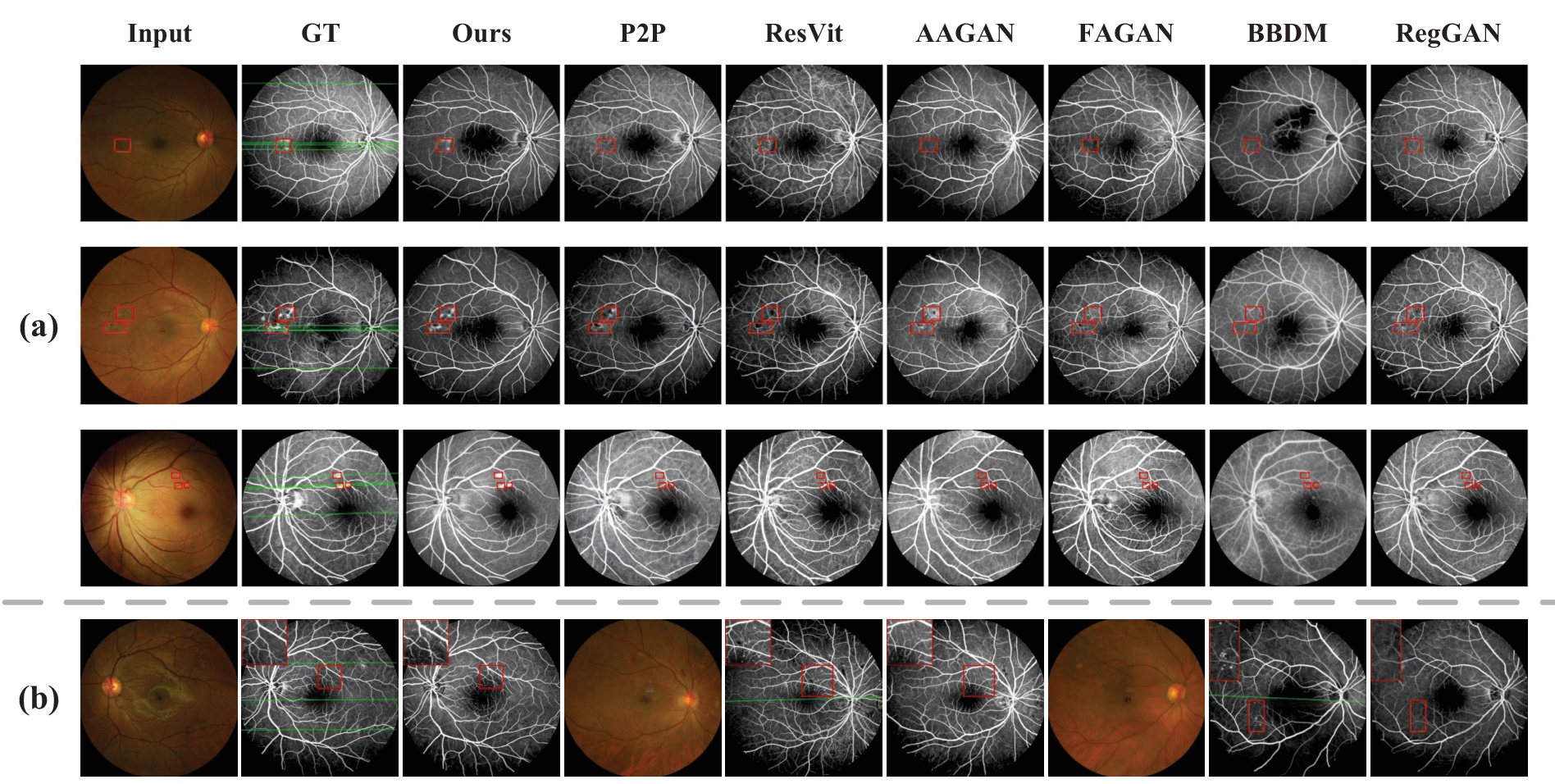}
    \caption{(a) Qualitative comparison of FFA synthesis results across different methods. (b) Failure cases and limitations of the proposed method. Areas in red boxes denote subtle leakage. Green lines indicate OCT scan trajectories. 
    % It can be observed that competing methods tend to miss small leakage spots, whereas our method preserves more faithful leakage patterns.
    }
    \label{fig:2}
\end{figure}

% As shown in Fig. \ref{fig:2}, competing methods often miss subtle leakage regions when conditioned only on CFP. In contrast, our OCT-guided model preserves finer leakage patterns and more complete vascular responses, reflecting improved pathology-aware synthesis. It can be observed that competing methods tend to miss small leakage spots, whereas our method preserves more faithful leakage patterns.
As shown in Fig. \ref{fig:2}, CFP-only methods often miss subtle leakage regions (red boxes), particularly small hyperfluorescent spots.
% As shown in Fig. \ref{fig:2}, methods conditioned solely on CFP often miss subtle leakage regions highlighted in the red boxes, particularly small hyperfluorescent spots. 
These approaches tend to produce over-smoothed responses, resulting in attenuated or absent leakage signals. 
In contrast, our OCT-guided model better preserves fine-grained leakage patterns, indicating improved pathology-aware synthesis. 
However, as shown in Fig. \ref{fig:2}(b), we also observe certain limitations. Vessel boundaries may appear blurred, and the model struggles to recover very small or widely distributed leakage regions, resulting in incomplete synthesis.

\begin{table}[t!]
    \caption{Ablation study on our proposed SACMF and TCMA modules. The best results are highlighted in \textbf{bold}.}
    \label{tab:ablation}
    {\small
        \centering

        \resizebox{0.98\linewidth}{!}{
            \begin{tabular}{c @{\hskip 1mm} c @{\hskip 3mm} cccc @{\hskip 3mm} cccc @{\hskip 2.5mm}}
                \toprule
                \toprule
                \multirow{2}{*}{SACMF} & \multirow{2}{*}{TCMA} & \multicolumn{4}{c}{Image Quality} & \multicolumn{3}{c}{Diagnosis (\%)} & \multicolumn{1}{c}{Semantic(\%)} \\ %
                \cmidrule(r){3-6} \cmidrule(r){7-9} \cmidrule(r){10-10} 
                & & PSNR$\uparrow$ & SSIM$\uparrow$ & FID$\downarrow$ & LPIPS$\downarrow$ & Pre$\uparrow$ & Rec$\uparrow$ & F1$\uparrow$ & Cos Sim$\uparrow$\\
                \midrule
                \ding{55} & \ding{55} & 15.36 & 0.4223 & 12.74 & 0.2135 & 46.58 & 45.44 & 45.84 & 76.58\\
                \ding{51}  & \ding{55} & 15.51 & 0.4349 & 10.42 & 0.2069 & 49.10 & 47.33 & 47.93 & 78.53\\
                \ding{51}  & \ding{51} & \bf{15.87} & \bf{0.4502} & \bf{10.05} & \bf{0.2009} & \bf{52.83} & \bf{50.61} & \bf{51.49} & \bf{79.13}\\	
                \toprule
                \toprule
            \end{tabular}
        }
        
        \vspace{1mm}
        \vspace{-5mm}
        %		\lbltbl{baselines}
    }
    
\end{table}

\subsubsection{Ablation Studies.}
% We conduct ablation studies to evaluate the contributions of SACMF and TCMA (Table \ref{tab:ablation}). Introducing SACMF yields notable improvements in both image quality and diagnosis. Specifically, FID decreases from 12.74 to 10.42 and LPIPS drops from 0.2135 to 0.2069, indicating better perceptual fidelity. Meanwhile, F1-score increases from 45.84 to 47.93, suggesting enhanced diagnostic utility. These gains demonstrate that SACMF effectively reduces cross-modal discrepancy by incorporating OCT structural cues.
% Further adding TCMA leads to consistent improvements across all metrics. PSNR and SSIM increase to 15.87 and 0.4502, while FID further decreases to 10.05. More importantly, F1-score rises to 51.49, and cosine similarity improves to 79.13, indicating stronger semantic alignment with real FFA. 
% The substantial improvement in diagnostic and semantic metrics confirms that TCMA enhances fine-grained cross-modal correspondence rather than merely optimizing pixel-level similarity. 
We conduct ablation studies to assess the contributions of SACMF and TCMA (Table \ref{tab:ablation}). Adding SACMF improves perceptual quality (FID 12.74→10.42, LPIPS 0.2135→0.2069) and increases F1-score from 45.84\% to 47.93\%, demonstrating effective incorporation of OCT structural cues. Further introducing TCMA yields consistent gains across all metrics, with PSNR/SSIM reaching 15.87/0.4502 and FID decreasing to 10.05. Notably, F1 rises to 51.49\% and cosine similarity to 79.13\%, indicating enhanced diagnostic performance and semantic alignment. These results suggest that TCMA enhances semantic-level cross-modal alignment beyond conventional pixel-wise reconstruction objectives.

\begin{figure}[t!]
    \centering
    \includegraphics[scale=0.19]{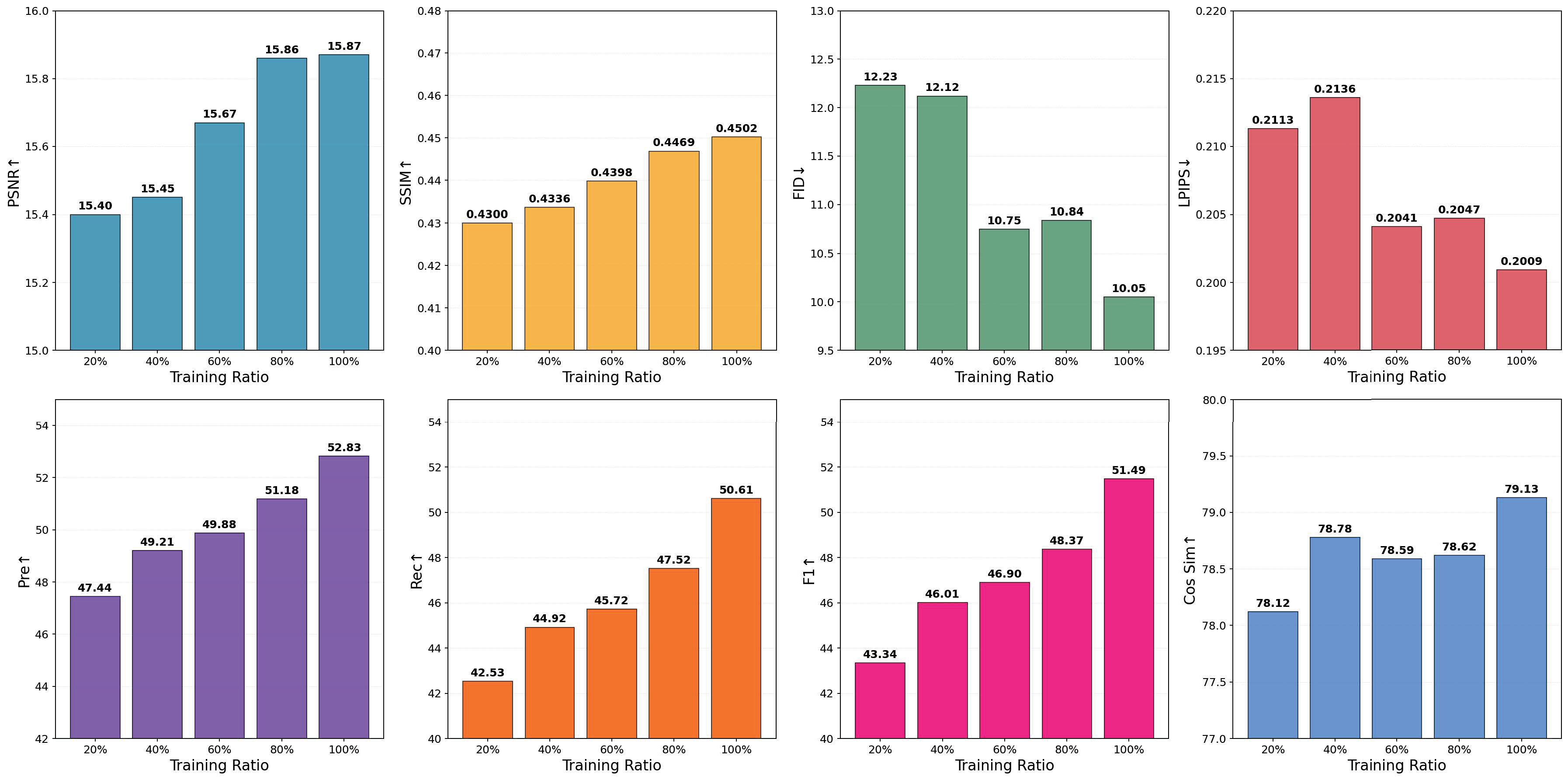}
    \label{fig:3}
    \caption{Impact of different OCT training ratios (20\%–100\%).}
\end{figure}

\subsubsection{Impact of OCT Training Ratio.}
We evaluate the impact of varying OCT training ratios (20\%–100\%) while keeping CFP unchanged. As shown in Fig.~\ref{fig:3}, as the OCT proportion increases, all metrics exhibit consistent improvement trends. 
These metric trajectories collectively confirm that denser OCT supervision strengthens structural guidance propagation and leads to more reliable cross-modal generation.

\section{Conclusion}
In this work, we present the first multi-modal CFP-to-FFA synthesis framework incorporating OCT as structural guidance. A tri-modal dataset with 3,676 paired CFP, FFA, and sparse OCT scans enables anatomically grounded learning. The proposed SACMF module projects depth-resolved OCT features onto the fundus plane for structural modulation, while TCMA enforces token-level CFP-FFA alignment. Experiments demonstrate superior image quality and discriminative representations for disease diagnosis. Integrating OCT alleviates the information gap in CFP-only translation, offering a practical direction for non-invasive angiography approximation in clinical settings.

\subsubsection{Acknowledgement.}
This work was partially supported by the National Natural Science Foundation of China (Grant No 62476054, Grant No 82401284, and Grant No 62576153).
\subsubsection{Disclosure of Interests.}
The authors have no competing interests to declare that are relevant to the content of this article.
%
% %% removed for anonymized MICCAI submission.
%    
%    % The following acknowledgement and disclaimer sections can be removed for the double-blind review process.  If and when your paper is accepted, reinsert the acknowledgement and the disclaimer clause in your final camera-ready version.
%    % IF you opted to include the acknowledgement and disclaimer sections, they will count towards the 8-page limit.

%\begin{credits}
%\subsubsection{\ackname} A bold run-in heading in small font size at the end of the paper is
%used for general acknowledgments, for example: This study was funded
%by X (grant number Y).
%
%\subsubsection{\discintname}
%It is now necessary to declare any competing interests or to specifically
%state that the authors have no competing interests. Please place the
%statement with a bold run-in heading in small font size beneath the
%(optional) acknowledgments\footnote{If EquinOCS, our proceedings submission
    %system, is used, then the disclaimer can be provided directly in the system.},
%for example: The authors have no competing interests to declare that are
%relevant to the content of this article. Or: Author A has received research
%grants from Company W. Author B has received a speaker honorarium from
%Company X and owns stock in Company Y. Author C is a member of committee Z.
%\end{credits}

%
% ---- Bibliography ----
%
% BibTeX users should specify bibliography style 'splncs04'.
% References will then be sorted and formatted in the correct style.
%
\bibliographystyle{splncs04}
\bibliography{Paper-1951.bib}
%\begin{thebibliography}{8}
%\bibitem{ref_article1}
%Author, F.: Article title. Journal \textbf{2}(5), 99--110 (2016)
%
%\bibitem{ref_lncs1}
%Author, F., Author, S.: Title of a proceedings paper. In: Editor,
%F., Editor, S. (eds.) CONFERENCE 2016, LNCS, vol. 9999, pp. 1--13.
%Springer, Heidelberg (2016). \doi{10.10007/1234567890}
%
%\bibitem{ref_book1}
%Author, F., Author, S., Author, T.: Book title. 2nd edn. Publisher,
%Location (1999)
%
%\bibitem{ref_proc1}
%Author, A.-B.: Contribution title. In: 9th International Proceedings
%on Proceedings, pp. 1--2. Publisher, Location (2010)
%
%\bibitem{ref_url1}
%LNCS Homepage, \url{http://www.springer.com/lncs}, last accessed 2023/10/25

%
%\end{thebibliography}
\end{document}